# Automated Knowledge Acquisition Pipeline for Disassembly Instructions: A Comparison of NLP and Generative AI Frameworks

Tim Schnieders[a]*, Tareq Maximilian Gabriel Ahrari[a], Oliver Petrovic[a], Christian Brecher[a]

[a]*Laboratory for Machine Tools and Production Engineering (WZL) RWTH Aachen University, Campus-Boulevard 30, 52074 Aachen, Germany*

* Corresponding author. Tel.: +49-241-8025434. *E-mail address:* t.schnieders@wzl.rwth-aachen.de

**Abstract**

The push towards circularity and Re-X principles presents a challenge to SMEs, particularly due to increasing product complexity that impedes the adoption of disassembly automation. This paper introduces and compares two frameworks for extracting disassembly instructions from unstructured information sources, such as disassembly manuals. Our goal is to transform natural language text into a structured, curated and machine-readable format. We implemented and tested a rule-based Natural Language Processing (NLP) framework against a generative AI framework leveraging LLMs. The performance of both approaches was compared using real-world disassembly manuals. This work provides the necessary data foundation for developing and automating future disassembly processes.

## 1. Introduction

The global transition toward a circular economy aims to replace the traditional linear "take-make-dispose" model by extending product lifecycles through Re-X processes, such as repairing, refurbishing, remanufacturing, and recycling [1–3]. While large companies can support circular technology development through internal research and development, Small and Medium-sized Enterprises (SMEs) often do not have the financial capacity to manage the disruptive transition to a circular business model and must depend on the availability of market technologies [4]. Furthermore, automated disassembly systems only act economically when processing large quantities of products, combined with the high upfront costs and time required for planning reverse logistics, this poses a significant barrier to the adoption of advanced disassembly automation for SMEs [2, 4].

One key enabler for a smart and digitally enabled circular economy is the implementation of the Digital Product Passport (DPP), which aims to provide consistent "track and trace" information on a product's origin, composition, and dismantling options [1, 5]. The DPP acts as a repository for diverse data categories, including guidelines and manuals for non-destructive disassembly [1]. However, the availability and accessibility of data for modeling remains a significant challenge, particularly for SMEs, creating a gap between theoretical DPP concepts and systems that can be readily used [6]. Current disassembly sequence planning still necessitates extensive manual effort, as operators must often search through available online documentation or rely on manually analyzing CAD models to infer priority relationships and disassembly steps [3]. To fully support Re-X ambitions and facilitate automated disassembly, there is a critical need for uniform data structures and automated data extraction methods to assist SMEs in obtaining and utilizing product data without excessive manual engineering efforts [1, 3].

This work focuses on working towards solving the issue of the manual high effort of finding the disassembly information within a given text and distilling the relevant aspects of a disassembly instruction such as needed tools, processes, positions, etc. into a structured depiction of different aspects to enable easy access for workers and inform automated disassembly processes with a structured, machine-readable format. The main contributions presented in this paper are:

- A framework utilizing a hybrid approach of Natural Language Processing and Large Language Models for the extraction of textual information from manuals into a structured format, specifically for disassembly processes
- A framework using solely the capabilities of Large Language Models to structure the textual information of disassembly processes into the same format as the hybrid approach.
- An approach to structure disassembly information and an approach to evaluate the performance of extraction frameworks on this structure.
- An experimental evaluation of the two presented frameworks on a newly created dataset of manuals and their disassembly instructions annotated with six categories of information elements.

## 2. State of the Art

Disassembly sequence planning (DSP) is a critical operation in the closed-loop supply chain, as it determines the order of operations required to retrieve valuable components or materials from end-of-life (EOL) products. Effective DSP can significantly reduce the environmental footprint of manufacturing by facilitating the reuse of high-value parts. Historically, research in DSP has focused on several key modeling methodologies to represent the relationships between components and the constraints of physical disassembly. The literature identifies Graph-based, AND/OR Graph, Petri Net (PN) and Matrix-based modeling as primary modelling categories used in modern disassembly research [7]. While these modeling methods have matured, they often require a structured representation of the product's assembly as an input. In many real-world scenarios, especially for SMEs dealing with diverse EOL products, the only available information is the textual manual. This creates a critical research gap: how to bridge the divide between natural language disassembly instructions and the structured input required for DSP models.

The rapid advancement of Natural Language Processing (NLP) and Large Language Models (LLMs) has presented a timely opportunity to address this gap. These technologies have become significantly more sophisticated in recent years, making them prime candidates for automating the extraction of technical knowledge. This paper introduces and evaluates a strategy to transform unstructured natural language from disassembly manuals into a curated, machine-readable format.

Previous research in complex fields like construction and medicine has actively compared traditional rule-based parsers against statistical machine learning models. These studies highlight that neither method performs flawlessly in isolation, as they each struggle with different aspects of unstructured data. For instance, while rule-based systems offer high precision for specific entities, they often lack the broad flexibility of learning-based models. Ultimately, the literature suggests that a hybrid approach, combining the strict logic of rules with the adaptability of smart models, consistently yields the best results [8–10].

In accordance with these findings this paper attempts to uncover the implications of applying rule-based parsers and statistical machine learning models to text-like representations of disassembly information and subsequently accurately and consistently extract the instructions within.

## 3. Implementation

To extract structured disassembly plans from unstructured manuals developed two extraction frameworks were developed, a rule-based NLP Parser and a GenAI extractor. The NLP Parser attempts to infer the correct labelling for each part of a disassembly instruction by utilizing a set of linguistic rules and analysis of the sentence structure**.** The GenAI extractor is given a system prompt which describes the process of determining the contents of each disassembly instruction.

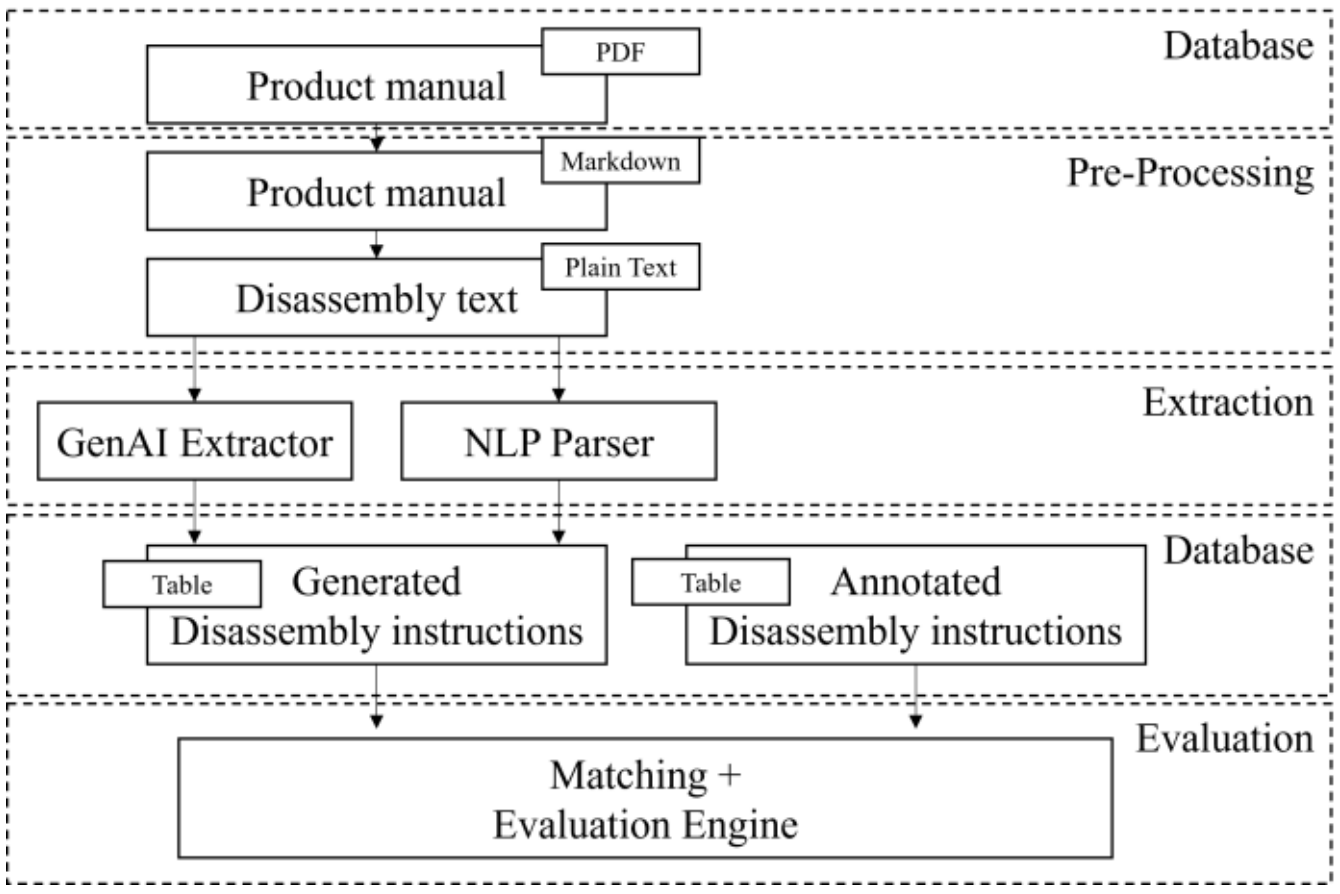

Fig. 1. Pipeline for extracting Disassembly instructions

### *3.1. Database*

The disassembly information can possess many forms such as a variety of video-, picture-, or text-data formats. Since this work focuses exclusively on text-data, the input for the two frameworks is constrained to text formats. The file formats to convey these texts are various. This paper focuses on only one, however files from any filetype that can have text extracted from it, can be utilized with this implementation and simply must replace the conversion into text with its own. A common format to convey text, particularly for manuals and handling instructions for products, is the PDF format. Therefore, this data format lends itself to a comparison of instruction extraction techniques.

### 3.2. Output format

The output format shall be structured, have easily identifiable categories and be able to be read into machines without any further language processing to facilitate an access point from which to go either towards full automation or a standardized instruction structure for humans. To address these needs the output has been defined as a table of regularly occurring elements of a disassembly instruction. These categories, called instruction categories in this research, include:

- **Action:** The action being taken, expressed as a verb lemma optionally combined with a particle or directional adverb. (e.g. remove, unscrew, disconnect, snap off etc.)
- **Object:** The component or part that the action is performed on. (e.g. screws, bolts, cover, panel, bearing etc.)
- **Tool:** The tool or instrument used to perform the action, if specified. (e.g. Phillips screwdriver, flat-head screwdriver, T-27 Torx wrench, hammer, brass drift punch etc.)
- **Count:** The quantity of objects involved, if explicitly stated. (e.g. 1, 2, 3, 4, 5, 6, 8, two, three, four etc.)
- **Condition:** A temporal, conditional, or simultaneous clause that constrains when or how the action is performed. (e.g. "after the system has completely shut down", "while pushing clip to the right", "once the spring pack is approximately halfway out" etc.)
- **Position:** The spatial relationship describing where the action takes place or where the object is located/moved from. (e.g. "from bin body", "on the display head", "in lower right corner", "into a soft-jaw vise" etc.)

The categories were determined by analyzing current research into disassembly planning methods, resulting in the categories Action; Object; Tool and Position. Reading through the available dataset revealed that these texts commonly contain other information that can potentially inform the disassembly process and help robotic or manual disassembly in defining the task, specifically Count and Condition.

### 3.3. Pre-Processing

The input data consists of unstructured disassembly manuals, containing elements like diagrams, tables, instructions and safety warnings. The goal for this work is to compare the NLP Parser and the GenAI Extractor specifically on text data. Thus, preprocessing of the available manuals is necessary.

First the data was transformed from PDF format into structured Markdown content using the Docling python library [11]. Second, the text was filtered for disassembly information from the Markdown using Gemini 2.5 Flash. This ensures that both approaches have a data input of plain text that is filtered to exclude other information, less relevant to disassembly planning, such as explanations of Legal Disclaimers, Product features, Cleaning and Maintenance as well as Warranty information.

### 3.4. Extraction

The rule-based **NLP Parser** was developed using the spaCy library, which provides part-of-speech tags, lemmas, and dependency parse trees for each sentence [12]. This information is used to fill in the instruction category. To that end, syntactic patterns were defined to extract information when grammatical rules are met. It achieves extraction by analyzing the sentence in two passes. In the first pass, the patterns "use <tool> to <verb>" are detected and the tool for later attachment is stored. In the second pass, all verbs whose dependency role within the spacy library are either *ROOT, conj, xcomp, advcl,* or *ccomp* are analyzed. For each verb, signaling a separate action, the categories contents are extracted in the following ways:

- **Action:** the verb lemma, extended with any verbal particle (e.g. "pull out") or directional adverb (e.g. "turn clockwise").
- **Object:** built from spaCy's noun chunks. During training, one recurring issue was the ambiguity of the word "and". Splitting sentences on "and" sometimes correctly separated two instructions, but in other cases incorrectly split objects that belong together. This issue was solved by avoiding splitting by solely detecting "and", and instead use spaCy's dependency parse to detect joined objects (e.g., "plug and packing") via the *.conjuncts* function, expanding them into separate instructions, sharing the same action.
- **Tool:** either found in first pass, otherwise from prepositional children such as "using", "with", "via", or "by" while skipping "followed by" which does not describe a tool.
- **Count:** identified by the numeric modifier in the object's subtree.
- **Condition:** extracted using a regex for temporal markers (e.g., "after", "before", "once"), with a fallback to dependencies that searches for prep labels.
- **Position:** selected from prepositional phrases attached to the object or verb. Prepositions that indicate tools or ownership (e.g. "of", "with", "by", "using") are excluded from position candidates.

The **GenAI Extractor** was developed based on an API that accesses Gemini 2.5 Flash. This model was chosen as an accessible API and promising results in preliminary test. The GenAI Extractor is separate from pre-processing and utilizes a different prompt for extracting instruction categories. We also provided examples to support Chain-of-Thought (CoT) reasoning [13, 14].

The CoT workflow includes the role assignment, "You are an AI assistant acting as technical advisor for disassembly manuals", as well as Step-by-step task instructions in the form of a numbered list telling the model how to think through the document:

- Carefully read through the markdown text of the manual.
- Identify and extract all disassembly steps, ignoring safety warnings, legal texts.
- Look for numbered lists and tables containing steps.
- Ensure that each disassembly step is unique; do not include duplicates.
- Always give the whole sentence for each step.
- Only include disassembly steps, not reassembly or assembly.
- Disassembly steps may include non-disassembly information that might look like assembly instructions but are still considered disassembly.

The CoT also includes explicit rules for irregularities such as double-count prevention, typo fixing, pronoun resolution, and tool hallucination prevention as well as category definitions with constraints on what each category means and what to include or exclude. Few-shot demonstrations were added to the prompt showing two standard examples and four edge cases showing input and expected output for the model.

The specific implementation is publicized under the name *"Neuro-symbolic-extraction-framework"* and can be found under the GitHub-link https://github.com/Tareqahr/Neuro-symbolic-extraction-framework.

## 4. Evaluation

To evaluate both frameworks, a custom evaluator was created that compares extraction outputs against a manually annotated target set. The work included for a human to manually annotate 741 instructions across 22 real-world disassembly manuals as ground truth. For each instruction, the six extraction categories were annotated: Action, Object, Count, Tool, Condition, and Position. The annotation treats each action as a separate instruction. This pertains specifically to occurrences of one sentence combining multiple instructions and for combined actions that apply the action to two different objects.

Since the two frameworks do not necessarily produce the same number of instructions as the ground truth, rudimentary index-based comparison is not possible. A Best-Match alignment, which is a greedy many-to-one matching where every prediction is matched to the target with the highest Jaccard similarity, was utilized. Multiple predictions may map to the same target [15]. This allows for extra instructions describing the disassembly steps in more detail instead of penalizing them as spurious, as shown in Fig. 2.

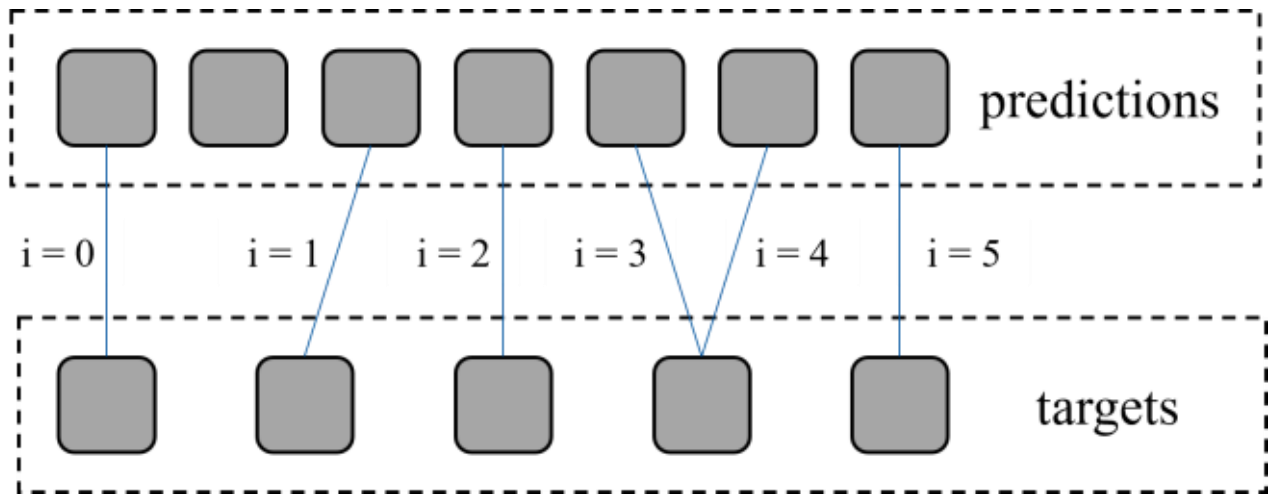

Fig. 2. Greedy many-to-one matching

To establish a coherent terminology human annotated results for the instructions are defined as targets with a subscript i for the index of each match found in the greedy many-to-one matching, and j for the index of the instruction category, $t_{i,j}$. The predictions of the models are annotated with the same subscript, $p_{i,j}$. Both these variables can consist of a set of multiple tokens.

For each matched pair, the six categories are compared individually. If both the predicted and target values are identical, the category scores 1.0. If one is empty and the other is not, the category scores 0.0. For partial matches, we award the category scores between 0.0 and 1.0 via Jaccard similarity, on token overlap, as shown in equation 1. Additionally, the per- category Jaccard similarity average across all matched rows is reported.

$$J(p_{i,j}, t_{i,j}) = \frac{|p_{i,j} \cap t_{i,j}|}{|p_{i,j} \cup t_{i,j}|} \quad (1)$$

The main metrics determined in this research are established in the literature and have been adapted to this use case. These include precision P (equation 2), recall R (equation 3), $F_1$ score (equation 4) and exact match EM (equation 5) [16].

$$P = \frac{TP}{TP+FP} \quad (2)$$

$$R = \frac{TP}{TP+FN} \quad (3)$$

$$F_1 = 2 \cdot \frac{P \cdot R}{P+R} \quad (4)$$

$$EM = \frac{\sum_{i=1}^{N} 1_{\{J(p_{i,j}, t_{i,j})=1\}}}{NumberOfAllTargets} \quad (5)$$

True positive *TP* is defined here as any match that has a Jaccard similarity for its full instruction of >=0.2. False positive *FP* are all predictions that could not be matched and thus are deemed as erroneous with a Jaccard similarity of <0.2. False negative *FN* are all instructions for whom there were no predictions with a Jaccard similarity of <0.2 and are deemed missed.

The experiments in this paper are carried out under the Windows 11 64-Bit operating system, the CPU is an 11th Gen Intel(R) Core(TM) i7-11800H, 2.30 GHz, with 15,7 GB usable RAM. The program is executed in Visual Studio Code on Python 3.14.3.

## 5. Results

The frameworks were evaluated on a corpus of real-world disassembly manuals comprising 741 manually annotated instructions. The manuals cover a range of products including consumer electronics (monitors, chargers), industrial machinery (wheelset bearings, differential assemblies), and household appliances. The results were reported using the Best-Match alignment, including coverage and per- category Jaccard similarity, as described in the evaluation section. Table 1 shows the main evaluation metrics for both systems.

Table 1. Evaluation metrics on all evaluated on all documents

| Predictions | NLP Parser (spaCy) | GenAI Extractor (Gemini 2.5 Flash) |
|---|---|---|
| *P* | 0.9137 | **0.9310** |
| *R* | **0.8718** | 0.8556 |
| $F_1$ | 0.8923 | 0.8917 |
| *EM* | **0.5013** | 0.3675 |

In Fig 3 the Jaccard similarity is averaged in each category as well as averaged over all categories.

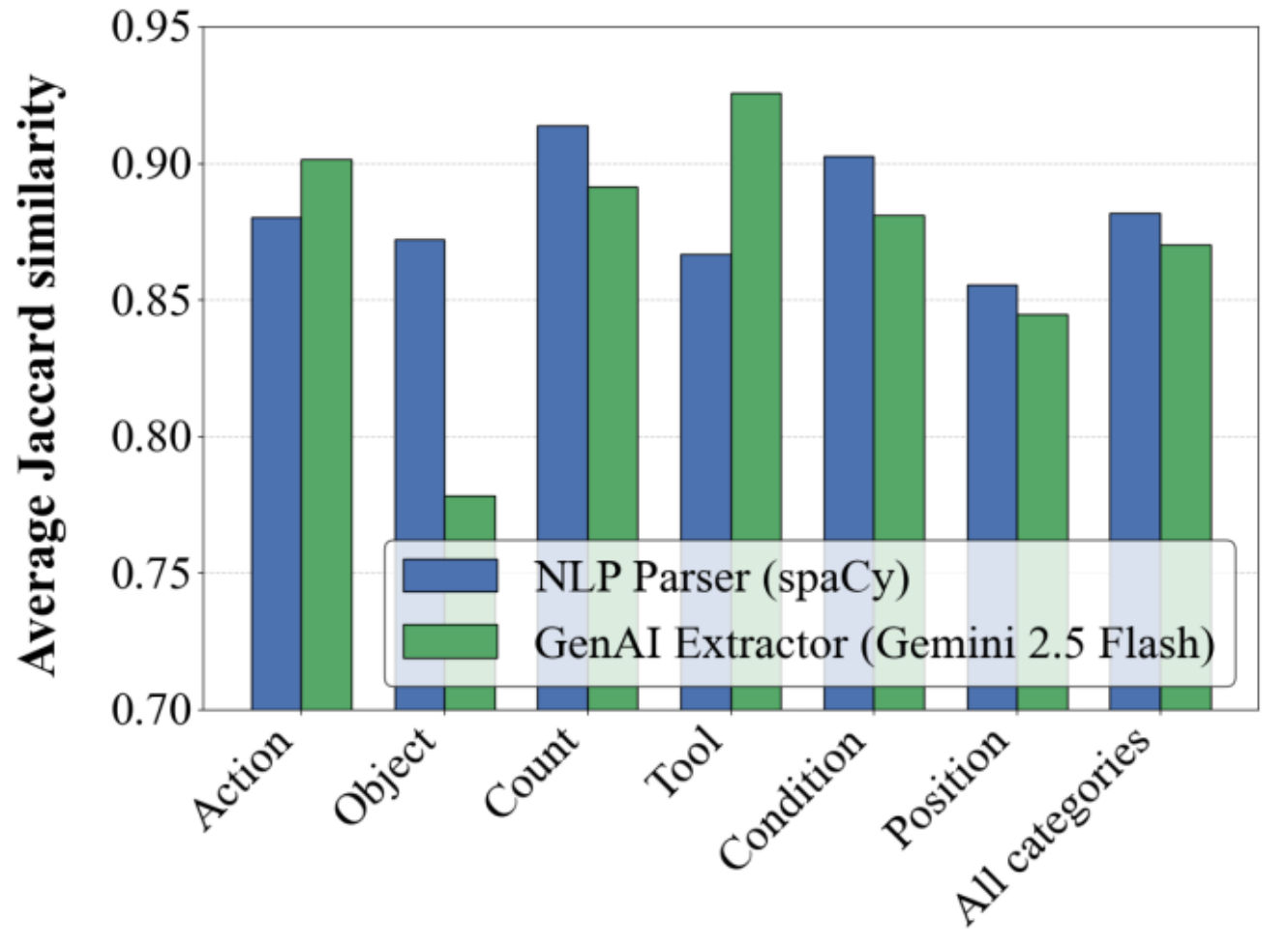

Fig. 3. Average Jaccard similarity by category

The results were additionally separated by complexity in Table 2. The necessary requirement to define a document as complex is for more than 15% of Tool, Condition, and Position categories in the target annotations to be non-empty. This applied to 15 documents in the dataset. Others were classified as simple (7 documents).

Table 2. Averaged Jaccard Similarity for all categories

| Document type | NLP Parser (spaCy) | GenAI Extractor (Gemini 2.5 Flash) |
|---|---|---|
| Short (≤ 20 instructions) | **0.8918** | 0.8566 |
| Long (> 20 instructions) | **0.8781** | 0.8752 |
| Simple | **0.9478** | 0.9423 |
| Complex | **0.8620** | 0.8487 |

Table 3 shows the processing time for each method as executed on the described hard and software. An important note is that the processing times on Gemini 2.5 Flash API can vary and thus depend on multiple factors such as traffic and network latency.

Table 3. Time consumption

| Metric | NLP Parser (spaCy) | GenAI Extractor (Gemini 2.5 Flash) |
|---|---|---|
| Model load (*s*) | 6.0 | **0.0** |
| Processing time (*s*) | **0.404** | 10.919 |
| Per instruction ($*10^{-3}$ *s*) | **33.6** | 909.9 |
| Throughput (*instructions/s*) | **29.7** | 1.1 |

Finally, the frameworks were evaluated on the number of instructions that they generated and how many of those matched with the annotated ground truth.

Table 4. Instruction Prediction for all evaluated documents

| Predictions | NLP Parser (spaCy) | GenAI Extractor (Gemini 2.5 Flash) |
|---|---|---|
| Matched | 687 | 671 |
| Unmatched | 20 | 10 |
| Total | 707 | 681 |
| Coverage | 0.972 | **0.985** |

## 6. Discussion

While still using the LLM Gemini 2.5 Flash to preprocess the documents, the difference between the implemented NLP Parser and the GenAI Extractor was demonstrated. Thus, the system still relies on the LLM to select what information will be processed but shows differing performance for parsing those steps into the predefined format.

As demonstrated in Table 1 the GenAI Extractor demonstrated a higher precision value, indicating that its predictions are less likely to be erroneous. However, the NLP Parser showed itself to extract a larger number of instructions. This tradeoff is reflected in the $F_1$ score, which has both frameworks at a similar performance. There is a difference in the number of exact matches that were achieved, on which the NLP Parser performed notably better. Indicating that, while the general instructions recovered from the text were similar, the NLP Parser was more consistent in extracting all annotated information on each instruction.

The average Jaccard similarity of both models also varies depending on what information category is extracted, as shown in Table 2. On the one hand, the GenAI Extractor outperforms the NLP Parser on the *Action* and *Tool* categories. This advantage likely stems from better handling of complex phrasal verbs, disambiguation of part-number words, with which *POS* tagging, used by the NLP Parser, struggles, and correct identification of the root verb in participial structures. While the NLP Parser occasionally fails to propagate the tool in "Use X to do Y" sentence structures, the GenAI Extractor demonstrates a superior extraction of tool-action relationships. On the other hand, the NLP Parser outperformed the GenAI Extractor in the categories *Object* and *Count*. The NLP Parser shows a favorable performance within the *Count* category presumably because it uses a targeted rule looking for *nummod* dependencies and *NUM POS* tags, whereas the GenAI extractor occasionally mistakes words like "both" for a numeric count. The advantage in the *Object* category might occur because the NLP parser deterministically splits conjoined objects into separate rows, which aligns with annotation procedure. The GenAI extractor tends to preserve natural language articles and sometimes merges multiple objects into a single row. While it certainly constitutes a difference in suitability to a specific annotation style, it does not inherently make the GenAI extractor less suitable for information extraction. The remaining categories carried similar results for both frameworks.

To make these categories machine readable, additional steps must be taken. This includes matching the contents of each category with a set of semantically pre-defined actions, objects, tools, conditions and positions by criteria such as cosine-similarity. This step is not discussed in this paper but constitutes a necessary step to enable robotic automation of disassembly.

## 7. Future Outlook

As the DPP becomes mandatory, the volume of available disassembly data will rise and so will the capabilities to implement Re-X strategies. Using web crawling and scraping services, this data can be supplemented with publicly accessible data.

To effectively utilize those capabilities, systems must be able to handle the amount and variety of this data. Future research should focus on the integration of automated extraction frameworks to build easily accessible knowledge base of disassembly instructions to address this challenge. This includes data formats that are not addressed in this paper, such as technical drawings and illustrations supplementing the disassembly information in product manuals. As well as CAD-data including 3D models and video demonstrations as they can be recorded by each company to fit their specific needs and are widely available for a range of consumer products on video sharing websites such as Youtube. Developing frameworks that can effectively process multimodal information beyond text will add crucial information to the disassembly process. A first step towards that goal is this research focusing exclusively on text-based instructions, the structured form of which can then also potentially be added to the DPP.

However, because no framework achieved an exact match in more than 50.13% of the instructions, there is a lot more research to do on how to build frameworks that can deliver more reliable results that extract all information into a structured output.

## Acknowledgements

Funded by the Deutsche Forschungsgemeinschaft (DFG, German Research Foundation) under Germany´s Excellence Strategy – EXC-2023 Internet of Production – 390621612.